\documentclass{article}

\usepackage{PRIMEarxiv}

\usepackage[utf8]{inputenc} % allow utf-8 input
\usepackage[T1]{fontenc}    % use 8-bit T1 fonts
\usepackage{hyperref}       % hyperlinks
\usepackage{url}            % simple URL typesetting
\usepackage{booktabs}       % professional-quality tables
\usepackage{amsfonts}       % blackboard math symbols
\usepackage{nicefrac}       % compact symbols for 1/2, etc.
\usepackage{microtype}      % microtypography
\usepackage{lipsum}
\usepackage{fancyhdr}       % header
\usepackage{graphicx}       % graphics
\graphicspath{{media/}}     % organize your images and other figures under media/ folder

\usepackage{eqparbox}
\usepackage{tabularx}
\usepackage{multirow}
\usepackage[labelformat=simple]{subcaption}

\DeclareCaptionLabelFormat{subcaptionlabel}{\normalfont(\textbf{#2}\normalfont)}
\title{Distract Your Attention: Multi-Head Cross Attention Network for Facial Expression Recognition
\thanks{\textit{\underline{Citation}}: 
\textbf{Wen, Z.; Lin, W.; Wang, T.; Xu, G. Distract Your Attention: Multi-Head Cross Attention Network for Facial Expression Recognition. Biomimetics 2023, 8, 199. https://doi.org/10.3390/biomimetics8020199}} 
}

\author{
  Zhengyao Wen, Wenzhong Lin, Tao Wang and Ge Xu \\
  Fujian Provincial Key Laboratory of Information Processing and Intelligent Control\\ College of Computer and Control Engineering, Minjiang University \\
  Fuzhou, China
}

\begin{document}
\maketitle

\begin{abstract}
This paper presents a novel facial expression recognition network, called Distract your Attention Network (DAN). Our method is based on two key observations in biological visual perception. Firstly, multiple facial expression classes share inherently similar underlying facial appearance, and their differences could be subtle. Secondly, facial expressions simultaneously exhibit themselves through multiple facial regions, and for recognition, a holistic approach by encoding high-order interactions among local features is required.
To address these issues, this work proposes DAN with three key components: Feature Clustering Network (FCN), Multi-head Attention Network (MAN), and Attention Fusion Network (AFN). Specifically, FCN extracts robust features by adopting a large-margin learning objective to maximize class separability. In addition, MAN instantiates a number of attention heads to simultaneously attend to multiple facial areas and build attention maps on these regions. Further, AFN distracts these attentions to multiple locations before fusing the feature maps to a comprehensive one. Extensive experiments on three public datasets (including AffectNet, RAF-DB, and SFEW 2.0) verified that the proposed method consistently achieves state-of-the-art facial expression recognition performance.
The DAN code is publicly available.
\end{abstract}

% keywords can be removed

\section{Introduction}
\label{sec:intro}
Facial expressions are direct and fundamental social signals in human communication~\cite{ekman1997face,darwin2015expression}. Along with other gestures, they convey important  nonverbal emotional cues in interpersonal relations. More importantly, vision-based
facial expression recognition has become a powerful sentiment analysis tool in a wide spectrum of practical applications.
For example, counseling psychologists assess a patient's condition and consider %Please check intended meaning is retained.
treatment plans by constantly observing facial expressions~\cite{fasel2003automatic}.
In retail sales, a customer’s facial expression data are used to determine whether a human sales assistant is needed~\cite{shergill2008computerized}.
Other significant application areas include social robots, e-learning, and facial expression~synthesis.

% Background for facial expression recognition
Facial expression recognition~(FER) is a technology that uses computers to automatically recognize facial expressions. As this research area matures, a number of large-scale facial expression datasets have emerged. In an early seminal work~\cite{ekman1971constants}, six prototypical emotional displays are postulated: angry (AN), disgust (DI), fear (FE), happy (HA), sad (SA), and surprise (SU), which are often referred to in the literature as \textit{{basic emotions}}. %MDPI: Please confirm if the italics should be retained.
 Recent FER datasets regard neutral (NE) or contempt (CO) as additional expression categories, expanding the number of facial expression categories to seven or eight.

% Challenge 1: intraclass vs interclass
In contrast to generic image classification, there are strong common features among different categories of facial expressions. Indeed, multiple expressions share inherently similar underlying facial appearance, and their differences could be less distinguishable. In computer vision, a common strategy to address this issue involves adopting a variant of the center loss~\cite{wen2016discriminative}. In this work, {a Feature Clustering Network~(FCN) is proposed}, which includes a simple and straightforward extension to the center loss, and it works well for optimizing both intra-class and inter-class variations. Unlike existing methods~\cite{cai2018island,li2018facial,farzaneh2021facial}, our method does not involve additional computations other than the variation of the cluster centers, {and it only requires a few hyper-parameters.}

\begin{figure}[]
\centering
\includegraphics[scale=0.98]{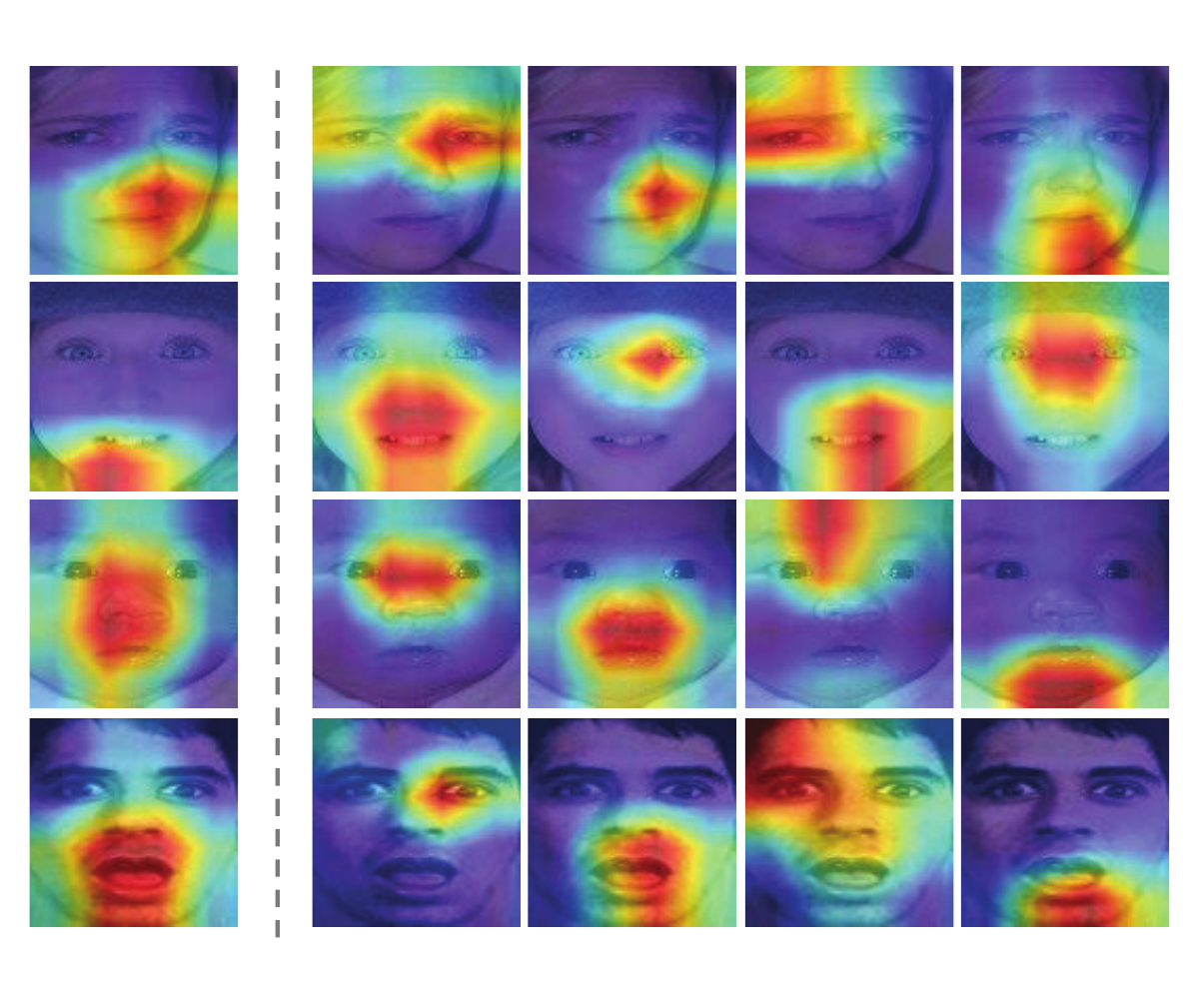}
\caption{{Comparing}  %MDPI: we moved the figure below where it is first mentioned, please confirm. 
the Grad CAM++ visualization of a single attention head model and our proposed DAN on the RAF-DB test set. The first column is obtained with DACL~{\cite{farzaneh2021facial},}   %MDPI: Please ensure that permission has been obtained and there is no copyright issue. If copyright is needed, please provide a citation in the following format: "Reprinted/adapted with permission from Ref. [XX]. Copyright year, copyright owner's name". More details on "Copyright and Licensing" are available via the following link: https://www.mdpi.com/ethics#10.
and the rest of the columns are generated by four attention heads from the proposed DAN model. Our method explicitly learns to attend to multiple local image regions for facial expression recognition.}
\label{fig:cam}

\end{figure}

% Challenge 2: local vs global
In addition, one unique aspect of FER lies in the delicate contention between capturing the subtle local variations and obtaining a unified holistic representation.
To attend to local details, some recent studies focus on attention mechanisms~\cite{farzaneh2021facial,fernandez2019feratt,li2020attention}, achieving promising results. Nonetheless, as shown in Figure~\ref{fig:cam}, it is difficult for a model with only a single attention head to concentrate on various parts of the face at the same time.
In fact, {it is our belief} that facial expression is simultaneously manifested in multiple parts of the face, such as eyebrows, eyes, nose, mouth, chin, etc.
Therefore, {this paper} proposes a Multi-head Attention Network (MAN) inspired by biological visual perception that instantiates a number of attention heads to attend to multiple facial areas. Our attention module implements both spatial and channel attentions, which allows for capturing higher-order interactions among local features while maintaining a manageable computational budget. Furthermore, {this paper} proposes an Attention Fusion Network (AFN) that ensures attention is drawn to multiple locations before their {fusion into a comprehensive feature vector for downstream classification}.

% a bit more details about our model
By integrating the above ideas, a novel facial expression recognition network, called Distract your Attention Network~(DAN), {is presented in this paper}. Our method implements multiple attention heads and ensures that they capture useful aspects of the facial expressions without overlapping. Concretely, {our work} proposes three sub-networks, including a Feature Clustering Network~(FCN), a Multi-head Attention Network~(MAN), and an Attention Fusion Network~(AFN). More specifically, {our method} first extracts and clusters the backbone feature embedding with FCN, where an affinity loss is applied to increase the inter-class distances while decreasing the intra-class distances. After that, an MAN is built to attend to multiple facial regions concurrently, where multiple attention heads that each include a spatial attention unit and a channel attention unit are adopted. Finally, {the output feature vectors from MAN} are fed to an AFN to output class scores. Specifically, this work designs a partition loss in AFN to force attention maps from the MAN to focus on different facial locations.
As shown in Figure~\ref{fig:cam}, a single attention module could only concentrate on one coarser image region, missing other important facial locations. On the contrary, our proposed DAN manages to simultaneously capture several vital facial regions.

% Main contributions
The main contributions of our work are summarized as follows:

\begin{itemize}

\item To maximize class separability, {this work} proposes a simple yet effective feature clustering strategy in FCN to simultaneously optimize intra-class variations and inter-class margins. 

\item {Our work} demonstrates that a single attention module cannot sufficiently capture all the subtle and complex appearance variations across different expressions. To address this issue, MAN and AFN {are proposed} to capture multiple non-overlapping local attentions and fuse them to encode higher-order interactions among local features.

\item The experimental results show that the proposed DAN method achieves an accuracy of 62.09\% on AffectNet-8, 65.69\% on AffectNet-7, 89.70\% on RAF-DB, and 53.18\% on SFEW 2.0, respectively, which represent the state of the art in facial expression recognition performance. 

\end{itemize}
%%%%%%%%%%%%%%%%%%%%%%%%%%%%%%%%%%%%%%%%%%

% Paper structure
The rest of the paper is organized as follows. Section~\ref{sec:related} reviews related literature in facial expression recognition with a particular focus on the attention mechanism and discriminative loss functions. Section~\ref{sec:method} describes the proposed method in detail. Section~\ref{sec:exp} then presents the experimental evaluation results followed by closing remarks in Section~\ref{sec:conclusion}.

\section{Related Work}
\label{sec:related}
Facial expression recognition is an image classification task involving accurately identifying the emotional state of humans. The earliest study on FER dates back to 1872, when Darwin first proposed the argument of consistency in expressions~\cite{darwin2015expression}. In 1971, Ekman and Friesen presented the six basic emotions~\cite{ekman1971constants}.
%expression labels of face (including angry, disgust, fear, happy, sad and surprise)
Later, the first facial expression recognition system~\cite{mase1991recognition} was proposed in 1991, which was based on optical flow. After that, the FER system has gradually matured and, in general, can be divided into three sub-processes: face detection, feature extraction, and expression classification~\cite{wu2012survey}. Recently, FER systems are benefiting from the rapid development of deep learning, and a unified neural network can be used to perform both feature extraction and expression classification. 

One of the most significant applications of FER is in human--computer interaction. By analyzing users’ facial expressions, computers can interpret human emotions and respond accordingly, providing more natural and intuitive user interfaces. This technology has already been implemented in various areas, such as gaming, virtual reality, and video conferencing. In the healthcare industry, facial expression recognition can be used to detect and diagnose various mental health disorders. It can also be used to monitor patients’ emotions and provide personalized treatment options. For example, {Ref.} \cite{app13053259} %mdpi: ref citation can not be in the beginning of the sentence, so we add a ``ref.'' before it, please confirm. 
take advantage of FER instead of medical devices to detect a people's health state. FER also has important applications in artistic research, {Ref.} \cite{s23052688} use FER to recognize facial expressions in opera performance scenes to help teams assess user satisfaction and use it to make adjustments. {Very recently,} \cite{dong2023recognizable} {propose a method for synthesizing recognizable face line portraits with controllable expression and high recognizability based on a triangle coordinate system.}

% attention mechanism enhances CNN model.
\noindent\textbf{{Attention mechanism.}}  %MDPI: Please confirm if the bold and noindent format should be retained.
Attention mechanism plays an important role in visual perception~\cite{rensink2000dynamic,corbetta2002control}. In particular, attention enables human beings to actively seek more valuable information in a complex scene. In recent years, there have been a plethora of studies attempting to introduce attention mechanisms into deep Convolutional Neural Network~(CNN) with success. For example,~\cite{hu2018squeeze} focus on the channel relationship of network features and propose a squeeze-and-excitation block to retain the most valuable channel information. {Ref.}~\cite{qin2021fcanet} {introduce frequency analysis to attention mechanism and present a new method termed FcaNet to compress information loss in scalar-based channel representations.} {Ref.}~\cite{li2019spatial} present a group-wise spatial attention module~(SGE) where the spatial-wise features are divided into multiple groups to learn a potential spatial connection. By leveraging the complementary nature between channel-wise and spatial-wise features, {Ref.}~\cite{woo2018cbam} propose the Convolutional Block Attention Module (CBAM) that sequentially connects a channel attention and spatial attention to obtain rich attention features. {Ref.}~\cite{hou2021coordinate} {propose a coordinate attention that embeds positional information into channel attention to generate spatially selective attention maps.} {Ref.}~\cite{misra2021rotate} {propose a new triplet attention that captures cross-dimensional interaction to efficiently build inter-dimensional dependencies.} Likewise, {Ref.}~\cite{fu2019dual} use a position attention module and a channel attention module in parallel to share the local-wise and global-wise features for scene segmentation task. Very recently, seminal works based on self-attention mechanisms have emerged, such as Swin-Transformer~\cite{liu2021swin}, DAB-DETR~\cite{liu2022dab}, {DaViT}~\cite{ding2022davit}, and {QFormer}~\cite{zhang2023vision}. These models have indeed demonstrated superior performance, but they also bring with them a huge number of model parameters and an expensive computational cost, which are unsuitable for lightweight applications. Inspired by these efforts, our method includes design of %Please check intended meaning is retained.
an attention head that sequentially cascades a spatial attention and a channel attention unit, which is both efficient and effective.

% attention mechanism used in FER field.
There are a few papers that introduce the above progress into FER. For example, {Ref.}~\cite{xie2019deep} apply region attention on the CNN backbone to enhance its power of capturing local information. {Ref.}~\cite{zhu2019discriminative} construct a bottom-up and top-down architecture to obtain low resolution attention features. In these papers, only a single attention head is used, which would generally lead to attention on a rough area of the face. In our work, however, multiple non-overlapping attention regions could be simultaneously activated to capture information from different local regions. This is particularly useful for FER, as we need to simultaneously attend to multiple regions (e.g., eyes, nose, mouth, forehead, and chin) to capture the subtle differences among emotion classes.

{Attention mechanisms have found applications in a wide range of areas in computer vision. For example,}~\cite{ning2021jwsaa} {propose a joint weak saliency and attention-aware model for person re-identification, in which saliency features are weakened to obtain more complete global features and diversified saliency features via attention diversity.} 
{Ref.}~\cite{chen2021image} {propose a reconstruction method based on attention mechanism to address the issue of low-frequency and high-frequency components being treated equally in image super-resolution.}
{Ref.}~\cite{wang2021dm3loc} {propose a multi-head self-attention approach for multi-label mRNA subcellular localization prediction and achieve excellent results.}

% several classical discriminative losses.
\noindent\textbf{{Discriminative loss.}} A discriminative loss function can strongly regulate the distribution of deep features. For example,~\cite{hadsell2006dimensionality} propose contrastive loss, which is an efficient loss function that maximizes class separability. In detail, a general Euclidean metric is used for features from the same class, but for diverse classes, the loss values will get close to a maximum margin. In addition,~{Ref.}~\cite{wen2016discriminative} present the center loss to learn a center distribution of each class and penalize the distances between deep features and their corresponding class centers. Differently, {Ref.}~\cite{liu2017sphereface} propose using an angle as a distance measure and introduce an angular softmax loss. That work is followed by a number of methods~\cite{liu2017learning,wang2018cosface,deng2019arcface} that improve the angular loss function.

% discriminative loss used in FER field.
In recent years, several studies demonstrate that discriminative loss functions could be well adapted to the FER task. {Ref.}~\cite{farzaneh2020discriminant} combine the advantages of center loss and softmax loss and propose a discriminative distribution-agnostic loss function. %Please check intended meaning is retained.
Concretely, a center loss aggregates the features of the same class into a cluster, and a softmax loss separates the adjacent classes. Similarly, {Ref.}~\cite{cai2018island} introduce a cosine metric based on center loss to increase the inter-class distance among different categories. Furthermore,~{Ref.}~\cite{farzaneh2021facial} propose an attentive center loss, which advocates learning the relationship of each class center for the center loss. However, all these loss functions bring in auxiliary parameters and computations. On the contrary, the affinity loss proposed in this paper is more simple and uses the internal relationship among class centers to increase the inter-class distance.

\section{Our Approach}
\label{sec:method}
% overview of our approach.
{This section} describes the proposed DAN model in detail.
In order to learn high-quality attentive features, our DAN is divided into three components: Feature Clustering Network~(FCN), Multi-head Attention Network~(MAN), and Attention Fusion Network~(AFN).
Firstly, the FCN accepts a batch of face images and outputs basic feature embedding with class discrimination abilities. Afterwards, the MAN is employed to learn diverse attention maps that capture several sectional facial expression regions. Then, these attention maps are explicitly trained to focus on different areas by the AFN. Finally, the AFN fuses features coming from all attention heads and gives a prediction for the expression category of the input images.

\begin{figure}[]
\centering
\includegraphics[scale=0.75]{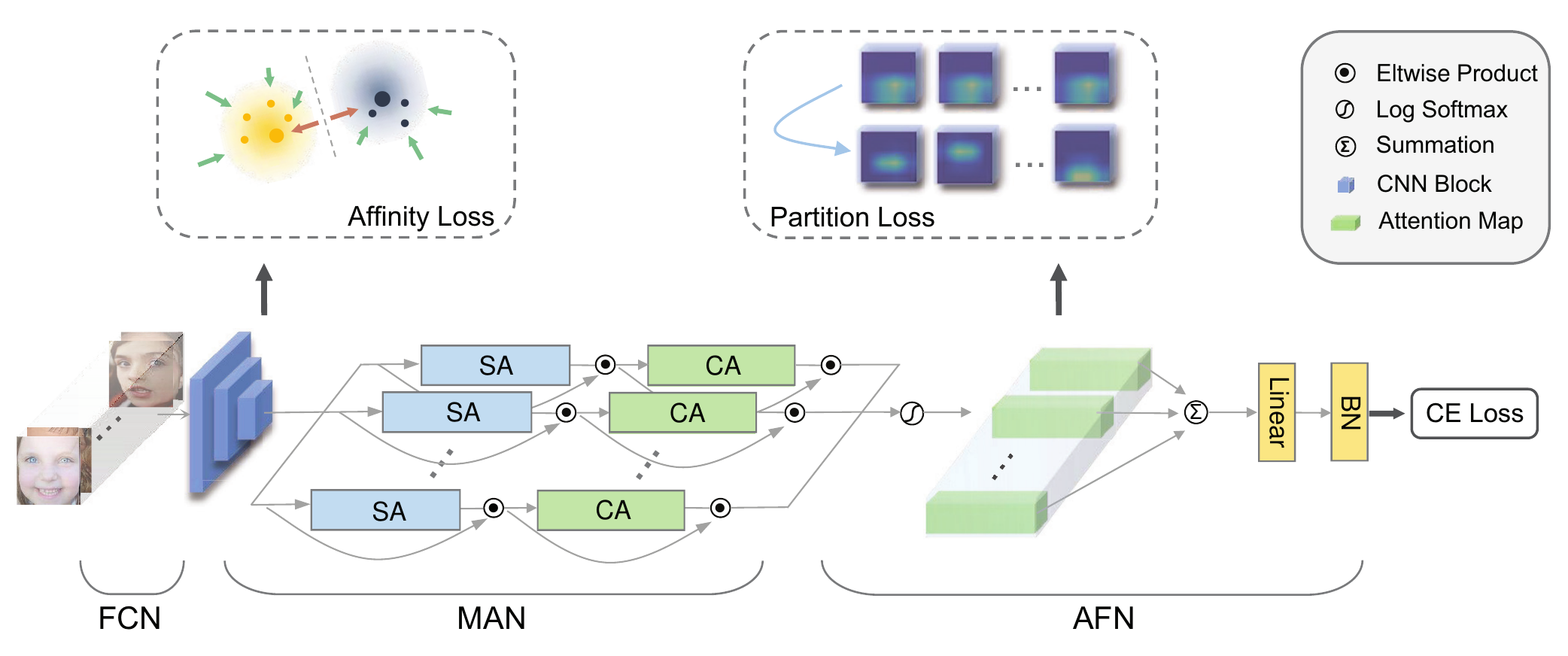}
\caption{{Overview}  %MDPI: we moved the figure below where it is first mentioned, please confirm. 
of our proposed DAN. The method is composed of three sub-networks. The backbone features are first extracted and clustered by a Feature Clustering Network~(FCN), where an affinity loss is applied to increase the inter-class margin and to reduce the intra-class variance. Next, a Multi-head Attention Network~(MAN) is built to attend to multiple facial regions concurrently by a series of  {Spatial Attention (SA) and Channel Attention (CA)} units. Finally, an Attention Fusion Network~(AFN) regulates the attention maps by enforcing variance among the attention feature vectors and outputs a class confidence.}
\label{fig:overview}
\end{figure}

In particular, the presented MAN contains a series of lightweight but effective attention heads. An attention head composes of a spatial attention unit and a channel attention unit in sequential order. Distinctively, the spatial attention unit involves convolution kernels of various sizes. A channel attention unit is connected to the end of the spatial attention unit to reinforce the attention maps by simulating an encoder--decoder structure. Both the spatial attention and the channel attention units are integrated back into the input features. The overall process of the proposed DAN is shown in Figure~\ref{fig:overview}.

\subsection{Feature Clustering Network~(FCN)}
Let us begin by introducing the FCN. Considering both performance and the number of parameters in our model, our method employs a residual network~\cite{he2016deep} as the backbone. As discussed earlier, different facial expressions may share similar underlying facial appearance. Therefore, this paper proposes a discriminative loss function, named affinity loss, to maximize class margins. Concretely, in each training step, our method encourages features to move closer to the class center to which they belong.
%% , with the class center being updated dynamically according the distribution of corresponding feature embeddings. 
At the same time, our method pushes centers of different classes apart in order to maintain good separability.

More formally, {suppose we have the \textit{i}-th input image feature $x_i\in \mathcal{X}$ with a class label $y_i\in \mathcal{Y}$ during training, where $\mathcal{X}$ is the input feature space and $\mathcal{Y}$ is the label space. Here, $i \in \{ 1 \dots M\}$ where $M$ is the number of images in the training set.} For the sake of simplicity, the output features of our backbone can be written as: 
\begin{equation}
    x_i^{'} = \mathcal{F}_r(w_r,x_i)
\end{equation}
\noindent where $\mathcal{F}_r$ represents the backbone network and $w_r$ denotes the network parameters of $\mathcal{F}_r$.

\noindent\textbf{{Affinity loss.}} %MDPI: Please confirm if the bold and noindent format should be retained.
The affinity loss is proposed to maximize the inter-class distance while minimizing the intra-class variation. By virtue of the affinity loss,
the backbone network can accurately cluster various features of facial expressions toward their respective class centers.
Compared to the standard center loss~\cite{wen2016discriminative}, our affinity loss advocates that it makes sense to expand the distance between class centers, because a larger distance between class centers further prevents overlap among classes. Formally, {given the class center matrix $c\in \mathcal{R}^{D \times|\mathcal{Y}|}$ wherein each column corresponds to the center of a specific class}, the affinity loss function can be written as follows:
%\textcolor{red}{TODO $x_i^'$ }
\begin{equation}
    \mathcal{L}_{af} = \frac{\sum_{i=1}^N||x_i^{'}-c_{y_i}||_2^{2}}{\sigma _c^2}
    \label{eqn:laf}
\end{equation}
\noindent {where $N$ is the number of images in a training batch, $D$ is the dimension of class centers, $c_{y_i}$ is the column vector in $c$ that corresponds to the ground-truth class of the $i$-th image}, and $\sigma _c$ indicates the standard deviation among class centers. {One might notice that $x_i^{'}$ is not necessarily a feature vector and could instead be a feature map of higher dimensions. In practice, following}~\cite{wen2016discriminative}{, a global average pooling is performed on $x_i^{'}$, and any singleton dimensions are then removed before applying Equation}~\ref{eqn:laf}, and we note that similar operations to flatten the feature maps may also work.

\begin{figure}[]
  \centering
  \begin{subfigure}[b]{0.48\textwidth}
    % \centering
    \includegraphics[width=\textwidth]{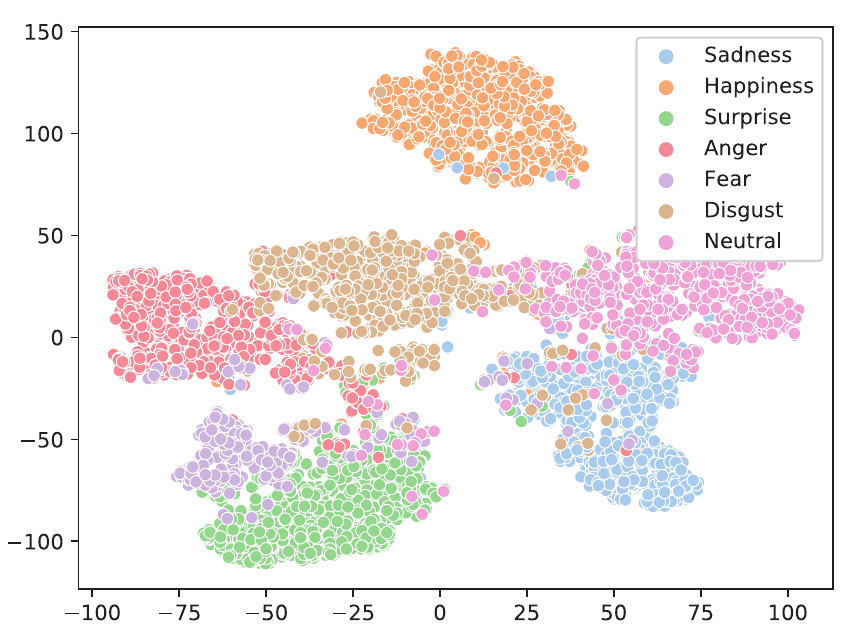}
    \captionsetup{justification=centering}
    \caption{}
    \vspace{2mm}
  \end{subfigure} 
  % \\
  \begin{subfigure}[b]{0.48\textwidth}
    % \centering
    \includegraphics[width=\textwidth]{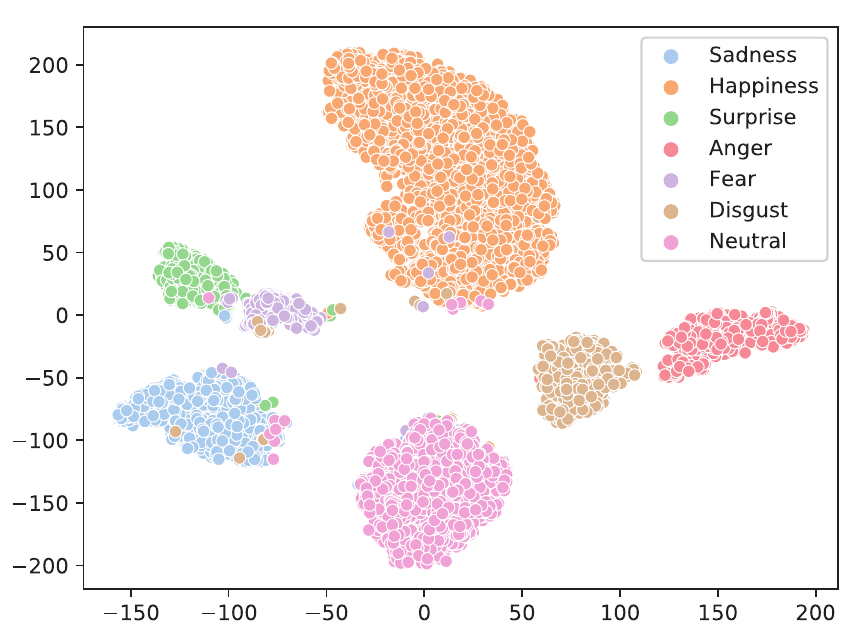}
    \captionsetup{justification=centering}
    \caption{}
    \vspace{2mm}
  \end{subfigure}
  \caption{The t-SNE visualization of features obtained with standard center loss (\textbf{a}) and the proposed affinity loss (\textbf{b}) on the RAF-DB dataset with our FCN, color-coded according to their class membership. It is clear that our affinity loss provides much better class separability by optimizing both the inter-class margins and the intra-class variations.}
  \label{fig:tsne}
\end{figure}

Despite its simplicity, our affinity loss is more effective than the standard center loss, as it encourages wider margins among classes as a result of pushing class centers further away from each other. This arguably completes the other half of the puzzle missing in the center loss, as the standard center loss only enforces tightness within classes but not sparsity among different classes. In particular, $\sigma _c$ can be readily available during training, as {class centers have already been computed}, and no additional hyper-parameters are needed for our affinity loss. Figure~\ref{fig:tsne} {presents} the t-SNE visualization of the features obtained by FCN with center loss and affinity loss, respectively. It is clear that by integrating the affinity loss, our FCN learns feature clusters of better quality, and the class margins are clearly wider among different classes.

\subsection{Multi-Head Attention Network~(MAN)}
\label{sec:man}
\begin{figure}[]
\centering
{\includegraphics[scale=1.5]{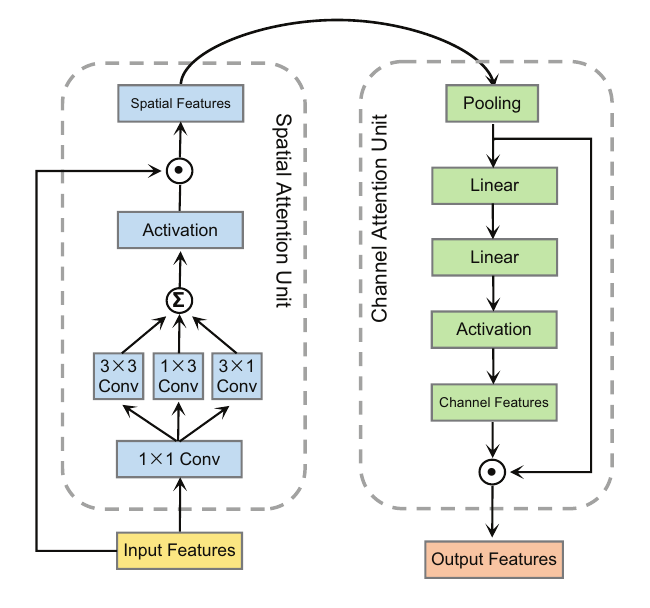}}
\caption{The structure of the proposed attention head, which consists of a spatial attention unit and a channel attention unit. The spatial attention unit and the channel attention unit work together to enable the attention head to selectively focus on relevant spatial locations and channel features.}
\label{fig:ca}

\end{figure}

As discussed earlier, a single attention head %Please check intended meaning is retained.
may be insufficient for effective FER, and the joint reasoning of multiple local regions is necessary. In this regard, our MAN contains a few parallel heads, which remain independent of each other. As shown in Figure~\ref{fig:ca}, an attention head is a combination of a spatial attention unit and a channel attention unit. The spatial attention unit receives the input features from the FCN and extracts the spatial features. Then, the channel attention unit accepts the spatial features as the input feature and extracts the channel features. Features from the two above dimensions are then combined into an attention vector. {In particular, it should be noted that all parallel heads are identical in terms of their sizes, differing only in parameter values.} %Please check intended meaning is retained.

More concretely, the left section of Figure \ref{fig:ca} illustrates the spatial attention unit, which consists of four convolutional layers and one activation function. In particular, our method constructs the $1 \times 1, 1 \times 3, 3 \times 1,$ and $ 3 \times 3$ convolution kernels to capture local features at multiple scales.
The channel attention unit shown in the right part consists of a global average pooling layer, two linear layers, and one activation function. In particular, our method takes advantage of having two linear layers to encode the channel information.

Formally, let $\mathcal{H}= \{H_1,  \ldots  , H_K \}$ be the spatial attention heads and $\mathcal{S}= \{s_1,  \ldots  , s_K \}$ be the output spatial attention maps, where $K$ is the number of attention heads. {Given the output features of our backbone $x^{'}$ (the subscript $i$ is omitted for notation brevity), the output of the $j$-th spatial attention unit can then be written as:}
\begin{equation}
    s_j = x^{'} \odot H_j(w_s, x^{'}), j\in\{1, \ldots ,K\}
\end{equation}
\noindent where $w_s$ are the network parameters of $H_j$, {and $\odot$ is the Hadamard product.}

Similarly, {assume $\mathcal{H}^{'}= \{H^{'}_1,  \ldots  , H^{'}_K \}$ are the channel attention heads and \linebreak $\mathcal{A}= \{a_1,  \ldots  , a_K \}$ are the final attention feature vectors of MAN. The $j$-th output $a_j$ can be written~as:}
\begin{equation}
    a_j = s_j \odot H^{'}_{j}(w_c, s_j), j\in\{1, \ldots ,K\}
\end{equation}
\noindent {where $w_c$ are the network parameters of $H_{j}^{'}$.}

% Further justify the novelty of MAN
The key benefits of the proposed MAN are two-fold. Firstly, our method incorporates spatial attention and channel attention sequentially for each attention head, so as to efficiently capture high-order interactions among features. More importantly, our method instantiates multiple attention heads so that the model can attend to different facial regions at the same time, which is not possible with a single attention head.

\subsection{Attention Fusion Network~(AFN)}
Furthermore, to guide the MAN in learning attention maps in an orchestrated fashion, this work proposes the AFN to further enhance the features learned by MAN.
Firstly, AFN scales the {attention feature vectors} by applying a log-softmax function to emphasize the most interesting region. After that, a partition loss is proposed to instruct the attention heads to focus on different crucial regions and avoid overlapping attentions. This would ensure the efficacy of multiple attention heads, making them attend to non-overlapping local regions. Lastly, the normalized {attention feature vectors} are merged into one, and we can then compute class confidence with a linear layer. 

% \noindent\textbf{Scaling attention maps.} To scale the attention maps from all heads, let $v$ be the output vectors of \textit{i}-th  head $a_i$, $v_p$ denote the \textit{p}-th vector of $a_i$, then $v^{'}$, the scaling result of log-softmax can be computed as:
% \begin{equation}
%     v^{'}_p = \log(\frac{e^{v_p}}{\sum_{q=1}^{k} e^{v_q}}),  p\in\{1,k\}
% \end{equation}

\noindent\textbf{{Feature scaling.}} {The final attention feature vectors $\mathcal{A}$ are firstly scaled using a log-softmax function along the dimension corresponding to the attention heads, so as to obtain a common metric for the features from different heads. More specifically, the output of MAN from Section}~\ref{sec:man} {gives the attention feature vectors $\mathcal{A} = \{ a_1, \ldots ,a_K\}$, and we assume that any of these vectors, e.g., the $j$-th vector $a_j$, is $L$-dimensional (e.g., $L=512$ in our experiments). In this case, the feature scaling result $v_j$ for $a_j$ can be given by:}
\begin{equation}
    v_{j}^{l} = \log\Big(\frac{\exp({a_{j}^{l}})}{\sum_{k=1}^{K} \exp(a_{k}^{l})}\Big),  l\in\{1, \ldots ,L\},  j\in\{1, \ldots ,K\}
\end{equation}

\noindent {where $a_{j}^{l}$ is the $l$-th element of $a_j$, and $v_j^{l}$ is the $l$-th element of $v_j$. After scaling, the features from different heads are further added up for the final classification, as shown in Figure~}\ref{fig:overview}.

\noindent\textbf{{Partition loss.}} This paper proposes a partition loss to direct the attention heads to focus on different facial feature regions so they will not collapse into a single attention. As visualized in the right part of Figure~\ref{fig:cam}, our partition loss successfully controls the heads of MAN to follow different facial areas without additional interventions. 
More specifically, we do so by maximizing the variance among attention vectors. In particular, our method considers $K$, the number of attention heads, as a parameter to adaptively adjust the descent speed of loss values. The MAN with a larger quantity of attention heads may generate more subtle areas of attention. Overall, we write our partition loss as follows:

\begin{equation}
    \mathcal{L}_{pt} = \frac{1}{NL} \sum_{i=1}^N\sum_{l=1}^L\log(1 + \frac{K}{\sigma_{i,l}^2})
\end{equation}
\noindent {where $N$ is the number of images in the current batch, $L$ is the dimension of the attention feature vector in each head, $\sigma_{i,l}^2$ denotes the variance of $v_j^l $ where
$j \in \{1, \ldots ,K\}$ when the \textit{i}-th image and the \textit{l}-th dimension are given.}

\subsection{Model Learning}
As shown above, our DAN model is comprised of three components: FCN, MAN, and AFN. In order to train this model, one has to consider the losses from all components (i.e., the affinity loss from FCN, the partition loss from AFN, and the cross-entropy loss for image classification) within a unified framework. Following the standard practice in deep learning, our method learns a final loss function by integrating all these loss functions:
\begin{equation}
    \mathcal{L} = \lambda_1\mathcal{L}_{af} + \lambda_2\mathcal{L}_{pt}  + \mathcal{L}_{cls}
\end{equation}

\noindent where $\lambda_1$, $\lambda_2$ are the weighting hyper-parameters for $\mathcal{L}_{af}$ and $\mathcal{L}_{pt}$. $\mathcal{L}_{cls}$ denotes the cross-entropy loss in image classification. {In our experiments, we empirically set both $\lambda_1$ and $\lambda_2$ to $1.0$ and note that the consistent performance gains we observe are not particularly sensitive to their values.}

\section{Experimental Evaluation}
\label{sec:exp}

{This section describes} our experimental evaluation results in detail. {In particular, the superiority of the proposed method is quantitatively verified} on three benchmark datasets: AffectNet, RAF-DB, and SFEW 2.0. This paper shows that the proposed method provides consistent performance improvements to a number of strong baselines. In addition, {it is demonstrated} that the various components of our model are all contributing to the final performance through ablation studies.

\subsection{Datasets}

\subsubsection{AffectNet} 
AffectNet~\cite{dhall2012collecting} is a large-scale database of facial expressions that contains two benchmark branches: AffectNet-7 and AffectNet-8 (with an additional category of ``contempt''). AffectNet-7 includes 287,401 images with seven classes, where all images are divided into 283,901 training samples and 3500 testing samples. Additionally, AffectNet-8 introduces the contempt data and expands the number of training and test samples to 287,568 and 4000, respectively.

\subsubsection{RAF-DB} 
RAF-DB~\cite{li2018reliable} is a real-world database with more than 29,670 facial images downloaded from the Internet. Seven basic and eleven compound emotion labels are provided for the dataset through manual labeling. There are 15,339  images in total for expression classification (including 12,271 training set and 3068 testing set images), each of which is aligned and cropped to a size of $100 \times 100$. 

\subsubsection{SFEW 2.0} 
\textls[-15]{SFEW 2.0~\cite{6130508} is the newest version for SFEW dataset in which each facial emotion is extracted from the static frames of the AFEW video database. It is divided into three sets of seven expression categories: train (958 samples), validate %Please check intended meaning is retained.
	(436 samples), and test (372~samples). Compared to the size of AffectNet and RAF-DB, SFEW 2.0 is light and compact.}

\subsection{Implementation Details}
On RAF-DB and AffectNet datasets, our work directly uses the official aligned image samples. For the SFEW 2.0 dataset, the facial images are manually aligned using the RetinaFace~\cite{deng2019retinaface} model. Input images are resized to $224\times 224$ pixels for each training and testing step on all datasets. A few basic random data augmentation methods~(such as horizontal flipping, random rotation, and erasing) are used selectively to prevent over-fitting. Moreover, a ResNet-18~\cite{he2016deep} model is adopted as the backbone of FCN in all experiments, and for a fair comparison, this work pre-trains the ResNet-18 model on the MS-Celeb-1M face recognition dataset. %TODO:Why?

Our experimental code is implemented with PyTorch, and the models are trained on a workstation with an NVIDIA Tesla P100 12GB GPU. {Our code is publicly available from} \url{https://github.com/yaoing/DAN}. For all tasks, models are trained for 40 epochs with {a uniform batch size of 256 (mainly due to GPU memory constraint)}, and the head number of attention in MAN is set to 4 by default.

More specifically, this work trains our model on the RAF-DB dataset using the SGD algorithm with an initial learning rate of 0.1. For the AffectNet-7 and AffectNet-8 datasets, the models are optimized by the ADAM algorithm with a smaller learning rate of 0.0001. Moreover, considering the inconsistent ratio of training set to testing set, this work introduces a dataset sampling strategy for the training step, i.e., upsampling the low-volume categories and downsampling the high-volume categories to obtain a more balanced dataset. %We set the batch size to 16 and a learning rate of 0.001 for the SFEW 2.0 dataset.

 \begin{table}[]
\caption{Performance comparison on the AffectNet-8 dataset.}
\label{table:affectnet8}
\newcolumntype{C}{>{\centering\arraybackslash}X}
\begin{tabularx}{\textwidth}{CC}
    \toprule
	\textbf{Methods}        & \textbf{Accuracy (\%)}\\ 
	\midrule
    PhaNet~\cite{liu2019pose}                 & 54.82      \\
    ESR-9~\cite{siqueira2020efficient}        & 59.30        \\
    RAN~\cite{wang2020region}             & 59.50            \\
    SCN~\cite{wang2020suppressing}            & 60.23         \\
    PSR~\cite{vo2020pyramid}                & 60.68         \\
    EfficientFace~\cite{zhao2021robust}  & 59.89            \\
    EfficientNet-B0 \cite{savchenko2021facial}  & 61.32         \\
    MViT~\cite{li2021mvit}            & 61.40                        \\
	\midrule
    ResNet-18                   & 56.84         \\
    DAN (ours)                  & \textbf{{62.09} %MDPI: Please confirm if the bold should be retained.
}         \\
    \bottomrule
\end{tabularx}

\end{table}

\begin{table}[]
\caption{Performance comparison on the AffectNet-7 dataset.}
\label{table:affectnet7}
\newcolumntype{C}{>{\centering\arraybackslash}X}
\begin{tabularx}{\textwidth}{CC}
    \toprule
	\textbf{Methods}        & \textbf{Accuracy (\%)} \\ 
	\midrule
	Separate-Loss~\cite{li2019separate}          & 58.89        \\
	FMPN~\cite{chen2019facial}                   & 61.25        \\
	LDL-ALSG~\cite{chen2020label}                 & 59.35         \\
	VGG-FACE~\cite{kollias2020deep}               & 60.00          \\
	OADN~\cite{ding2020occlusion}                 & 61.89         \\
    DDA-Loss~\cite{farzaneh2020discriminant}       & 62.34       \\   
    EfficientFace~\cite{zhao2021robust}         &  63.70    \\
    MViT~\cite{li2021mvit}                     & 64.57      \\
	\midrule
    ResNet-18                 & 56.97         \\
    DAN (ours)                  & \textbf{{65.69} %MDPI: Please confirm if the bold should be retained.
}         \\
    \bottomrule
\end{tabularx}
\end{table}

\begin{table}[]
\caption{Performance comparison on the RAF-DB dataset.}
\label{table:rafdb}
\newcolumntype{C}{>{\centering\arraybackslash}X}
\begin{tabularx}{\textwidth}{CC}
    \toprule
	\textbf{Methods}        & \textbf{Accuracy (\%)}\\ 
	\midrule
	Separate-Loss~\cite{li2019separate}          & 86.38   \\
    DDA-Loss~\cite{farzaneh2020discriminant}       & 86.9              \\
    SCN~\cite{wang2020suppressing}            & 87.03              \\
    PSR~\cite{vo2020pyramid}            & 88.98        \\
    DACL~\cite{farzaneh2021facial}           & 87.78             \\
    IF-GAN~\cite{cai2021identity}         & 88.33              \\
    EfficientFace~\cite{zhao2021robust}  & 88.36             \\
    MViT~\cite{li2021mvit}            & 88.62          \\
	\midrule
	ResNet-18         & 86.25   \\
	DAN (ours)       & \textbf{{89.70} %MDPI: Please confirm if the bold should be retained.
}                     \\
    \bottomrule
\end{tabularx}
\end{table}

\begin{table}[]
\caption{Performance comparison on the SFEW 2.0 dataset.}
\label{table:sfew}
\newcolumntype{C}{>{\centering\arraybackslash}X}
\begin{tabularx}{\textwidth}{CC}
    \toprule
	\textbf{Methods}        & \textbf{Accuracy (\%)} \\ 
	\midrule
	IACNN~\cite{meng2017identity}          & 50.98      \\
    DLP-CNN~\cite{li2018reliable}        & 51.05                 \\
    Island Loss~\cite{cai2018island}    & 52.52              \\
    TDTLN~\cite{yan2019cross}           & 53.10            \\
    RAN~\cite{wang2020region}            & 54.19              \\
    LDL-ALSG~\cite{chen2020label}        & 56.50            \\
    ViT + SE~\cite{aouayeb2021learning}  & 54.29           \\
    FaceCaps~\cite{wu2021facecaps}       & \textbf{{58.50} %MDPI: Please confirm if the bold should be retained.
}   \\
    \midrule
	Baseline (ResNet-18)                  & 51.29          \\
	DAN (ours)       & 53.18               \\

    \bottomrule
\end{tabularx}
\end{table}
\begin{table}[]
\caption{Ablation studies for the loss functions in FCN and AFN on the RAF-DB dataset. The proposed affinity loss and partition loss both provide superior performance. {Note that the cross-entropy loss is used in all cases.}}
\newcolumntype{C}{>{\centering\arraybackslash}X}
\begin{tabularx}{\textwidth}{CCC}
\toprule
\textbf{Task} &   \textbf{Methods}   & \textbf{Accuracy (\%) }\\ 
\midrule
\multirow{3}{*}{FCN}
& -                      & 88.17    \\
& center loss             & 88.91    \\
& affinity loss        &    89.70 \\
% \bottomrule

\midrule
\multirow{2}{*}{AFN} 
 & -                     &    88.20     \\
& partition loss         &    89.70 \\
\bottomrule
\end{tabularx}

\label{table:fcn_afn}

\end{table}

\subsection{Quantitative Performance Comparisons}
This paper presents the quantitative performance comparison results in Tables~\ref{table:affectnet8}--\ref{table:sfew} for AffectNet, RAF-DB, and SFEW 2.0. %Please check intended meaning is retained.
Our proposed method achieves an accuracy of $62.09\%$ on AffectNet-8 and an accuracy of $65.69\%$ on AffectNet-7, which are both superior to existing methods. Comparing for the RAF-DB dataset, DAN acquires $89.70\%$ in accuracy, which is state of the art. Our method, although not the best, also obtains a competitive accuracy of $53.18\%$ on the more compact SFEW 2.0 dataset. This is perhaps because the multi-head attention would require larger datasets for learning to ensure %Please check intended meaning is retained.
an effective model. Overall, the results above clearly demonstrate that the proposed method is highly competitive and the components in our method are effective across multiple datasets.

\subsection{Ablation Studies and Computational Complexity}

\subsubsection{Effects of Loss Functions for FCN and AFN}
In order to verify the efficacy of the individual loss functions proposed in our method, this paper now performs ablation studies on the RAF-DB dataset to demonstrate that our affinity loss and partition loss are indeed contributing to model performance. Specifically, this work evaluates the influence of using different loss functions for FCN and AFN separately in Table~\ref{table:fcn_afn}. In FCN, our proposed affinity loss provides a considerable performance improvement to the standard center loss. Similarly, the partition loss plays a crucial role in the performance of AFN. These results demonstrate that the affinity loss and the partition loss both contribute to the superior performance of our model.

\subsubsection{Number of Attention Heads}
\begin{figure}[]
\centering
{\includegraphics[scale=0.44]{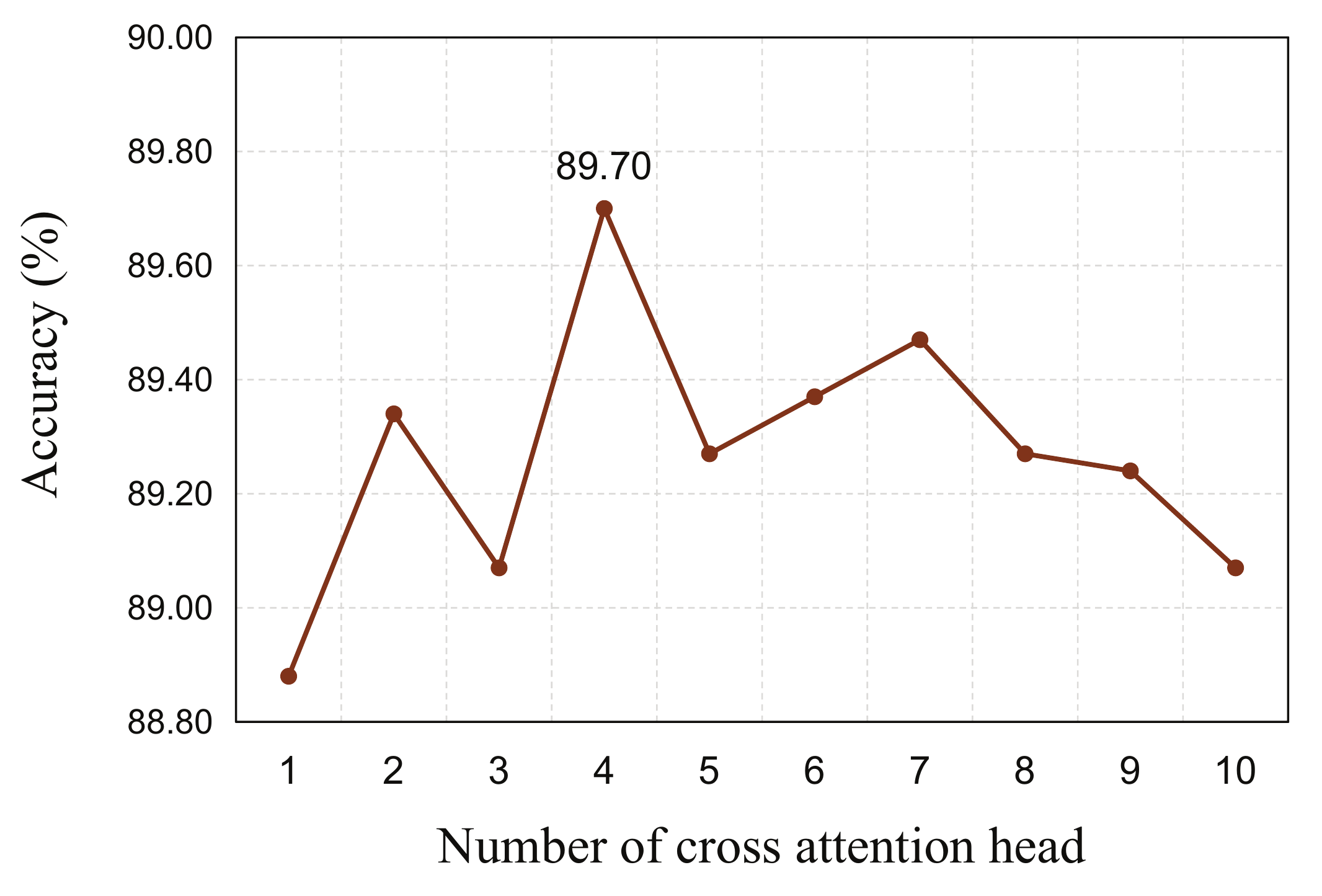}}
\caption{Ablation studies for the changing number of attention heads of MAN on the RAF-DB dataset. It is clear that our empirically chosen model with four heads is superior to the model with a single attention head.}
\label{fig:num_head}
\end{figure}

 In addition, the number of attention heads obviously affects the performance of our model. Figure~\ref{fig:num_head} shows the accuracy results with a changing number of attention heads on the RAF-DB dataset. It is evident that our proposed MAN structures are superior to a single attention module. Moreover, using four attention heads maximizes the performance gain. Therefore, four attention heads are used throughout the experiments in this paper.

\begin{table}[]
\caption{Comparison based on model size and inference time. Our method provides competitive performance while maintaining a manageable computational cost.}
\newcolumntype{C}{>{\centering\arraybackslash}X}
\begin{tabularx}{\textwidth}{CCC}
    \toprule
	\textbf{Methods}              & \textbf{Params (M)}        & \textbf{FLOPs (G)} \\ 
    \midrule
    EfficientFace~\cite{zhao2021robust}   & 1.28    & 0.15    \\
    SCN~\cite{wang2020suppressing}    & 11.18    & 1.82    \\
	RAN~\cite{wang2020region}    & 11.19    & 14.55    \\
	PSR\cite{vo2020pyramid}         & 20.24   & 10.12  \\
	DACL~\cite{farzaneh2021facial}   & 103.04      &1.91      \\
	\midrule
    ResNet 18         & 11.69    & 1.82  \\
	DAN (ours)        &    19.72     &2.23   \\
    \bottomrule
\end{tabularx}
\label{table:flops}

\end{table}

\begin{figure}[]
    \begin{subfigure}[b]{0.485\textwidth}
        \centering
        \includegraphics[width=\textwidth]{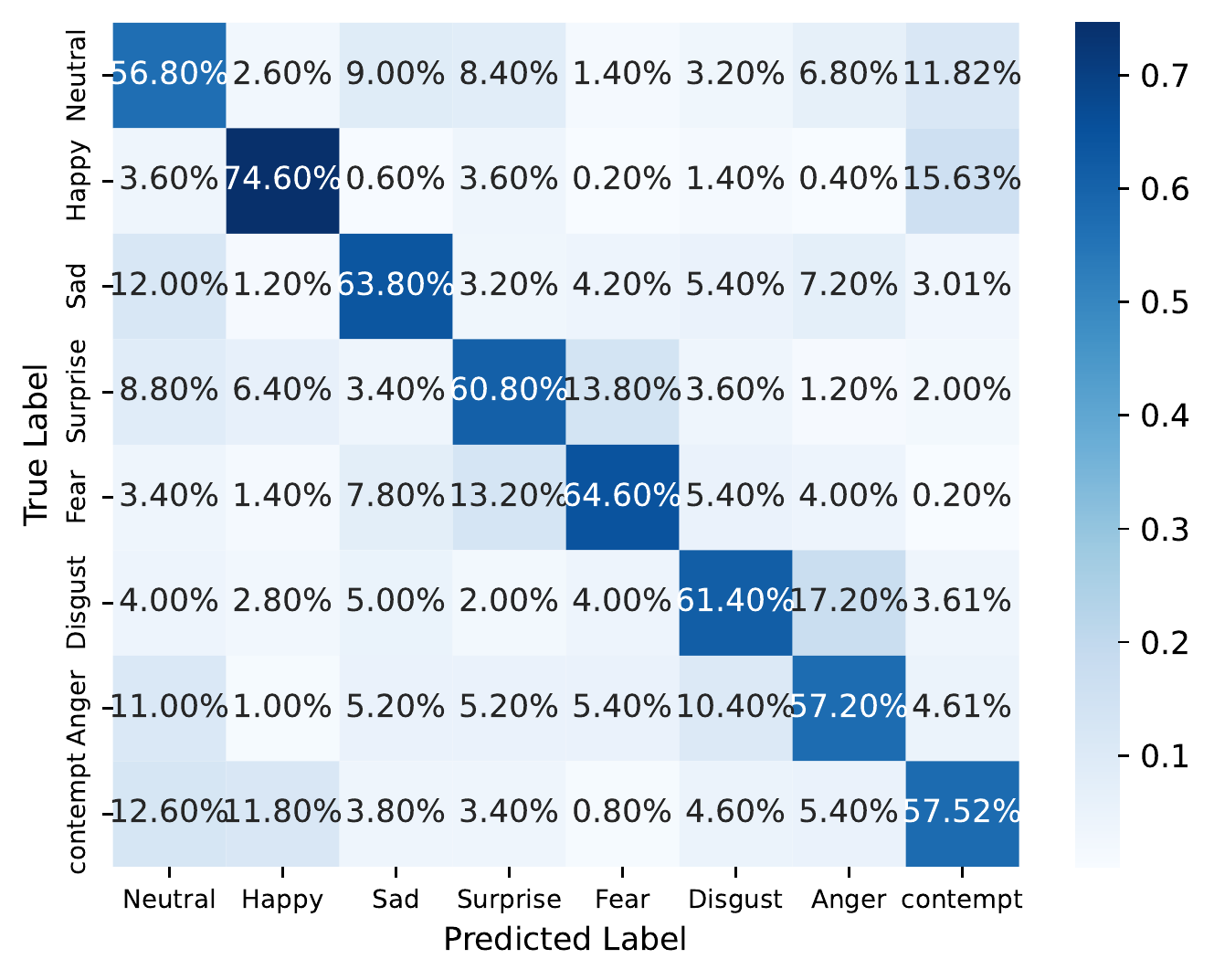}
        \captionsetup{justification=centering}
        \caption[]%
        {{\small Confusion matrix on AffectNet-8}}    
        \label{fig:mean and std of net14}
    \end{subfigure}
    \begin{subfigure}[b]{0.485\textwidth}  
        \centering 
        \includegraphics[width=\textwidth]{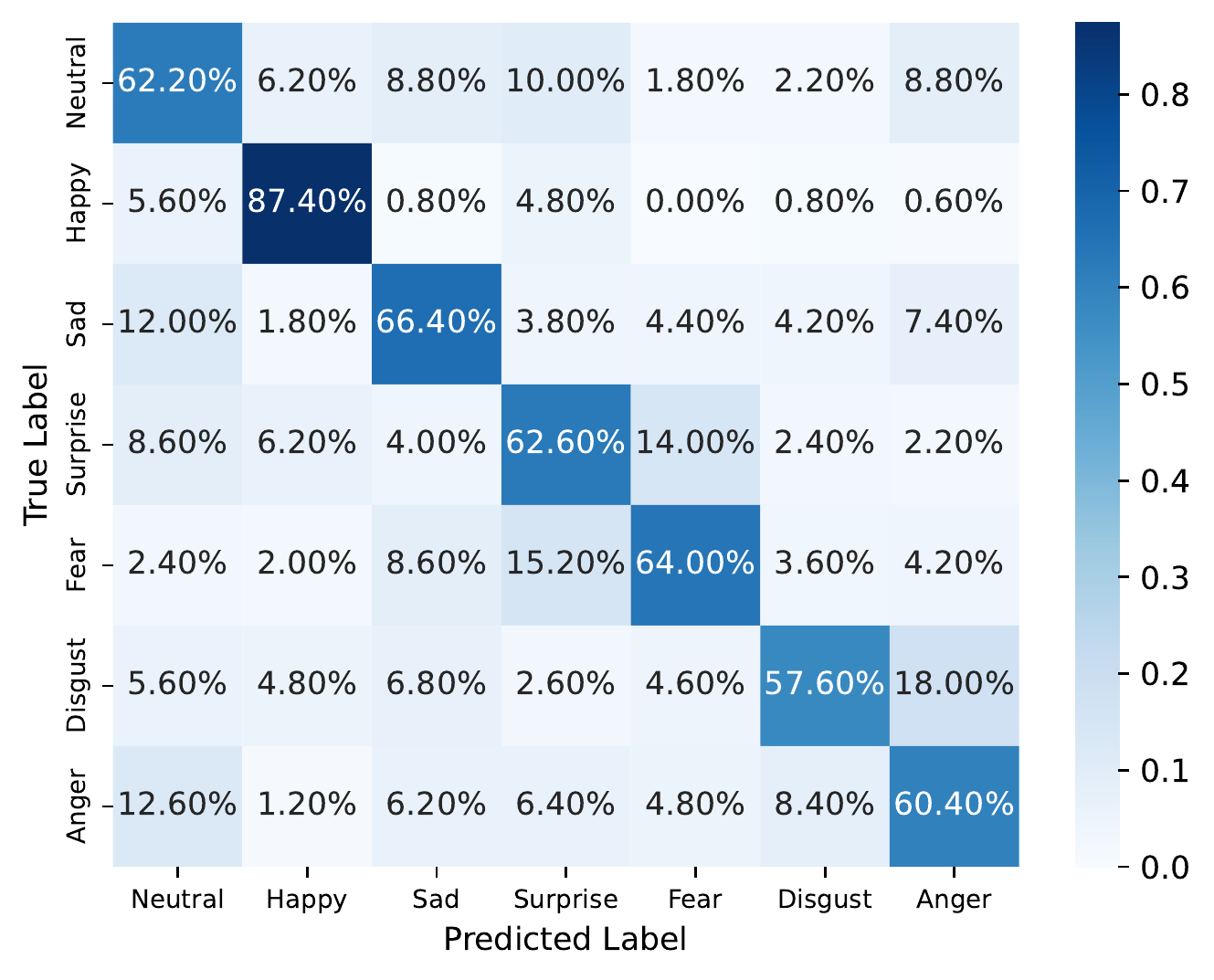}
        \captionsetup{justification=centering}
        \caption[]%
        {{\small Confusion matrix on AffectNet-7}}    
        \label{fig:mean and std of net24}
    \end{subfigure}
  \vspace{6pt} \\
    \begin{subfigure}[b]{0.485\textwidth}   
        \centering 
        \includegraphics[width=\textwidth]{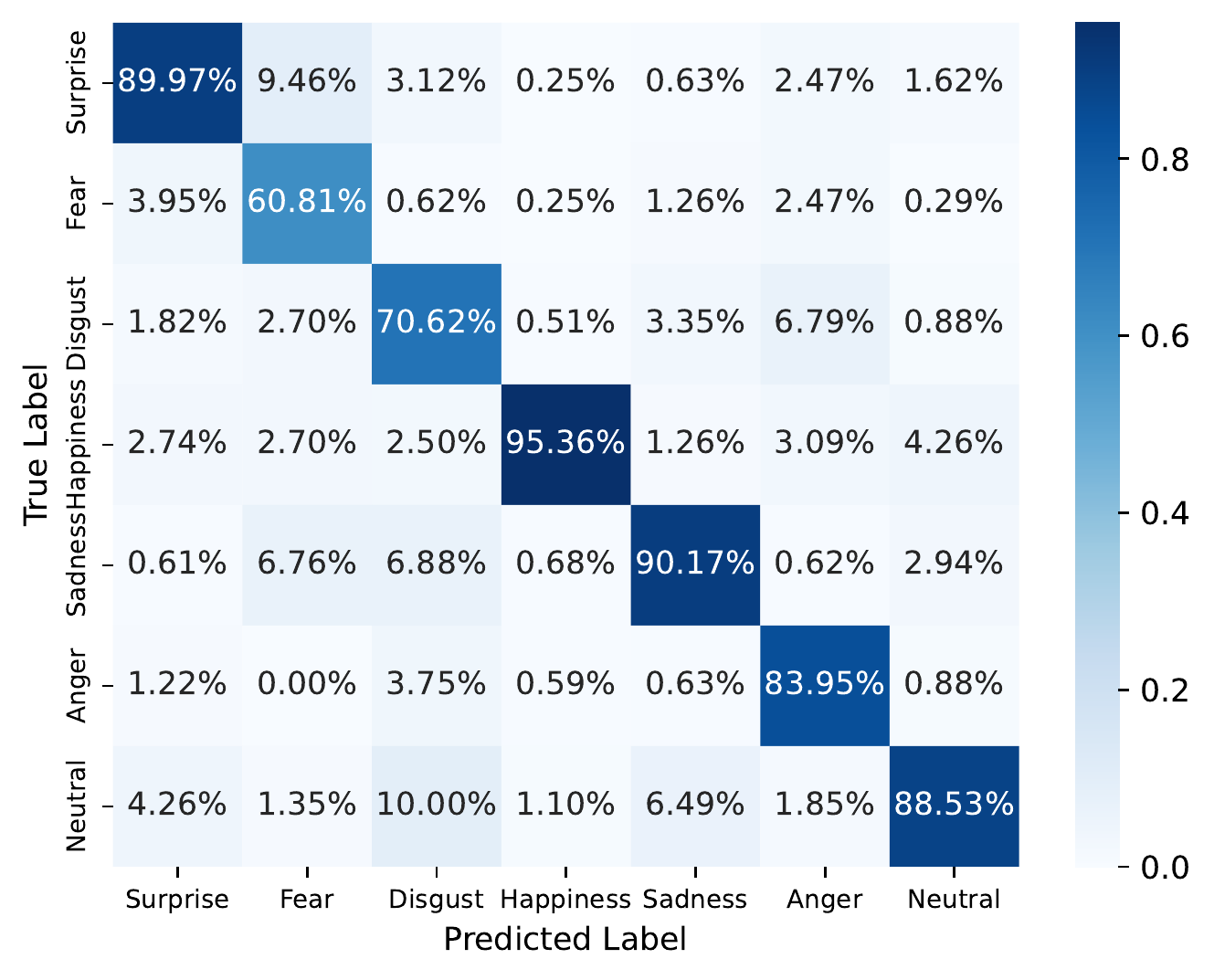}
        \captionsetup{justification=centering}
        \caption[]%
        {{\small Confusion matrix on RAF-DB}}    
        \label{fig:mean and std of net34}
    \end{subfigure}
    \begin{subfigure}[b]{0.485\textwidth}   
        \centering 
        \includegraphics[width=\textwidth]{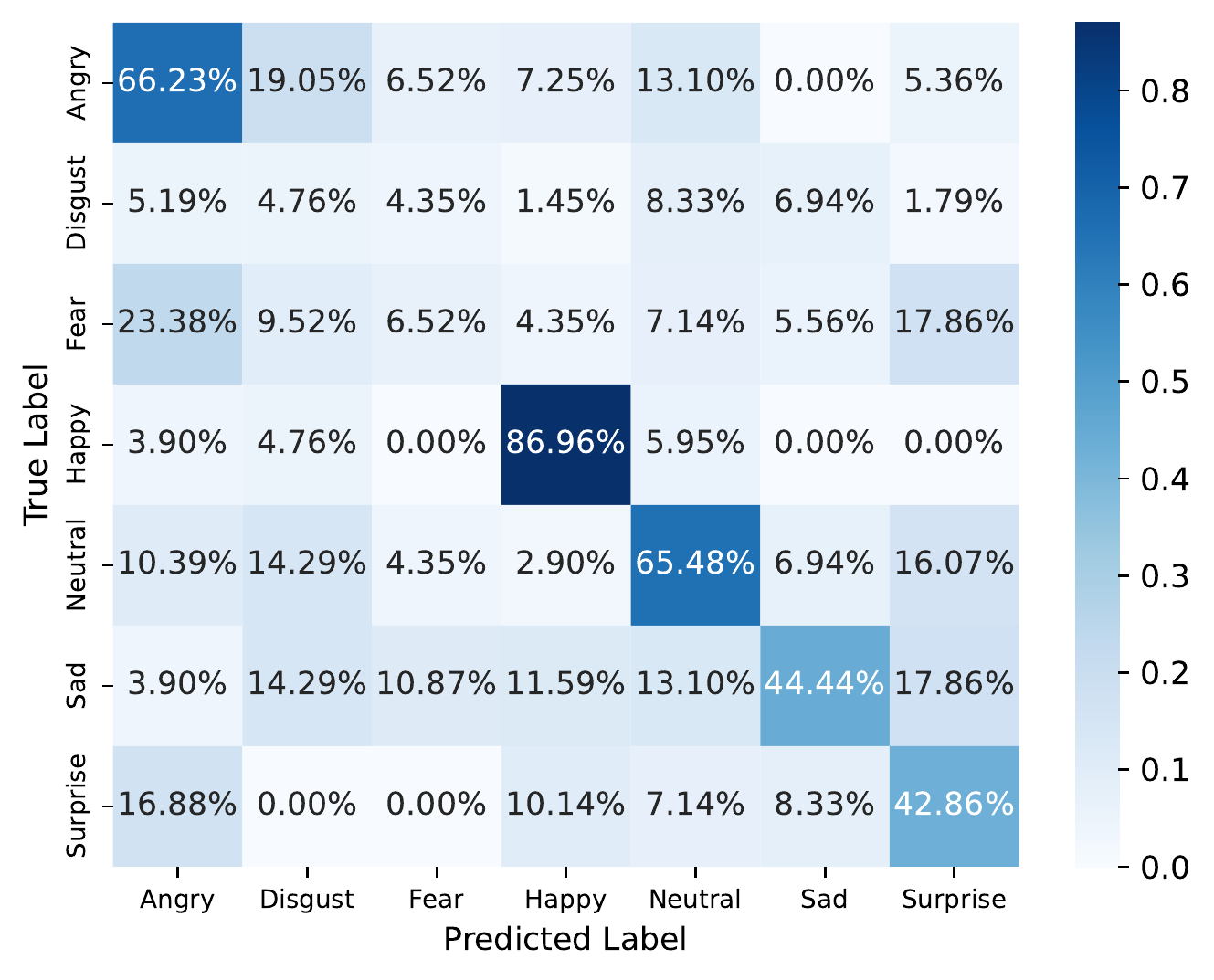}
        \captionsetup{justification=centering}
        \caption[]%
        {{\small Confusion matrix on SFEW 2.0}}    
        \label{fig:mean and std of net44}
    \end{subfigure} \vspace{6pt}
    \caption{ Confusion matrix on AffectNet-8, AffectNet-7, RAF-DB, and SFEW 2.0, respectively.} 
    \label{fig:cm}
\end{figure}

\subsubsection{Computational Complexity}
 Table~\ref{table:flops} compares our method against state-of-the-art methods in terms of model size and inference time. Our DAN with four attention heads uses 19.72~M and 2.23~G of parameters and FLOPs, respectively, which corresponds to moderate resource consumption, to achieve its state-of-the-art facial expression recognition performance.

\subsection{Confusion Matrix}
In order to better understand the performance of our model on each facial expression category, Figure~\ref{fig:cm} presents the confusion matrices on each dataset we evaluated. More specifically, it is obvious that the ``Happy'' class is the easiest on both Affect-8 and Affect-7, followed by ``Sad'', ``Surprise'', and ``Fear''. On RAF-DB, our DAN model has a high accuracy in ``Happiness'', ``Sadness'', ``Surprise'', ``Neutral'', and ``Anger'' classes. On SFEW 2.0, our method performs relatively well on the ``Happy'' and ``Neutral'' %Please check intended meaning is retained.
classes, while the ``Disgust'' and ``Fear'' classes are very challenging. Possible reasons for the large gaps in the performance include the appearance similarities among facial expression categories as well as the skewed class distribution in the training dataset. It is therefore obvious that to further improve the performance of our method in the future, extra efforts should be made to avoid class confusion on the more difficult classes such as ``Disgust'', ``Fear'', and ``Anger''.

\subsection{Precision--Recall Analysis}
\begin{figure}[]
\centering
{\includegraphics[width=8.5cm]{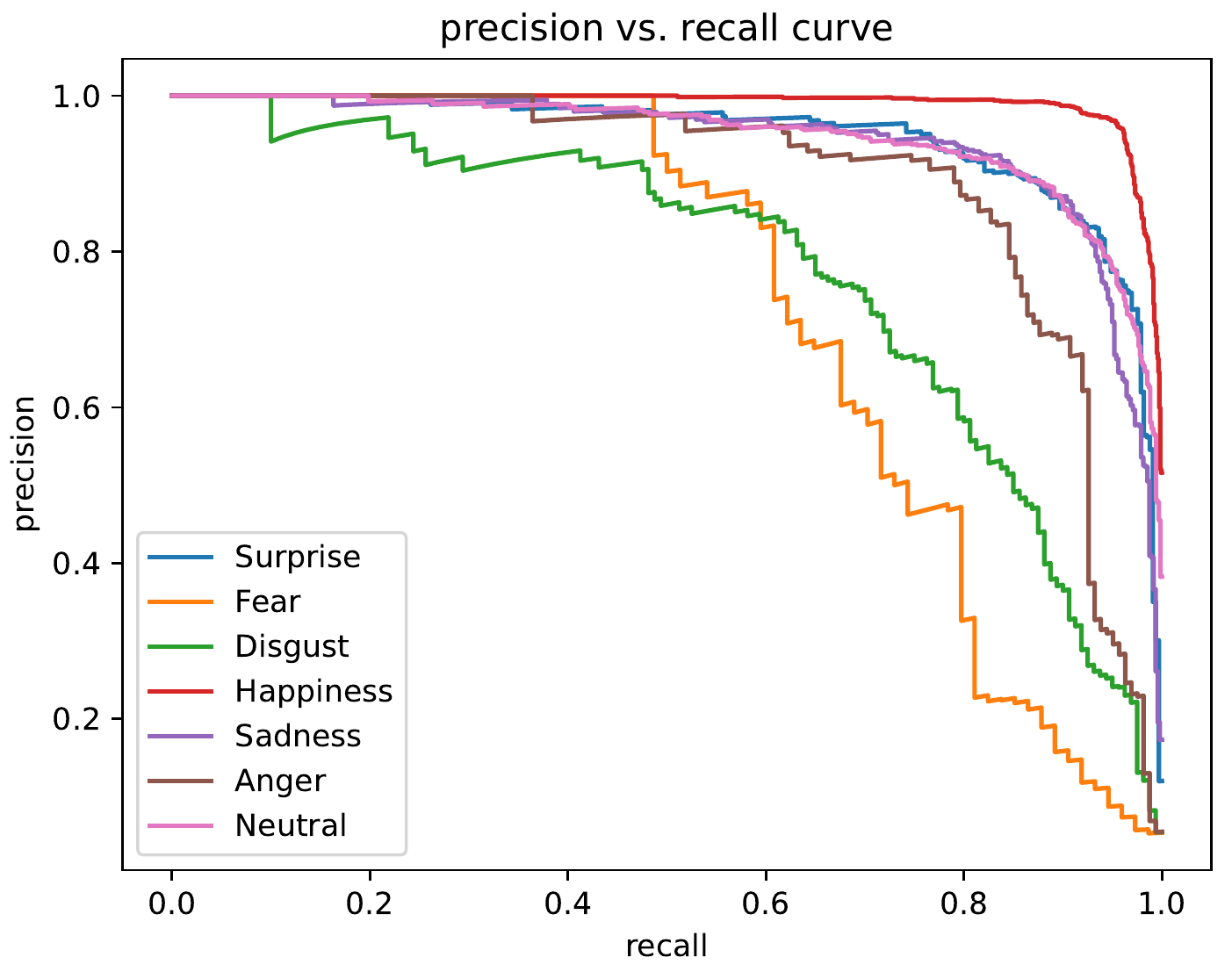}}
\caption{The per-class precision--recall curves on RAF-DB. Our method provides near-perfect precision for most classes in the low recall regime. Classes such as ``Happiness'', ``Neutral'', ``Surprise'', and ``Sadness'' are relatively easy, while the disgust and fear classes are the most challenging.}
\label{fig:prc}

\end{figure}

Finally, Figure~\ref{fig:prc} presents the per-class precision--recall curves on the RAF-DB dataset. It is evident that our approach achieves near-perfect performance in ``Happiness'' class, and also performs well in ``Neutral'', ``Surprise'', ``Sadness'', and ``Anger'' classes. Overall, our method provides over 80\% precision at 50\% recall for all classes.

\section{Conclusions}
\label{sec:conclusion}
This paper presents a robust method for facial expression recognition that consists of three novel sub-networks including the Feature Clustering Network~(FCN), the Multi-head Attention Network~(MAN), and the Attention Fusion Network~(AFN). Specifically, the FCN learns to maximize class separability for backbone facial expression features, the MAN captures multiple diverse attentions, and the AFN penalizes overlapping attentions and fuses the learned features.
Experimental results on three benchmark datasets demonstrate the superiority of our method for FER. 
%For the SFEW 2.0 dataset, the improvement is not conspicuous enough, suggesting that our DAN takes large-scale samples to %
%learn how to cluster the underlying features and distract the attentions to crucial regions.
It is our hope that the exploration into feature clustering and learning multiple diverse attentions would provide insights for future research in facial expression recognition and other related vision tasks.

%Bibliography
\bibliographystyle{unsrt}  
\bibliography{references}

\begin{thebibliography}{10}

\bibitem{ekman1997face}
Paul Ekman and Erika~L Rosenberg.
\newblock {\em What the face reveals: Basic and applied studies of spontaneous
  expression using the Facial Action Coding System (FACS)}.
\newblock Oxford University Press, USA, 1997.

\bibitem{darwin2015expression}
Charles Darwin.
\newblock {\em The expression of the emotions in man and animals}.
\newblock University of Chicago press, 2015.

\bibitem{fasel2003automatic}
Beat Fasel and Juergen Luettin.
\newblock Automatic facial expression analysis: a survey.
\newblock {\em Pattern recognition}, 36(1):259--275, 2003.

\bibitem{shergill2008computerized}
Gurvinder~Singh Shergill, Abdolhossein Sarrafzadeh, Olaf Diegel, and Aruna
  Shekar.
\newblock Computerized sales assistants: the application of computer technology
  to measure consumer interest-a conceptual framework.
\newblock 2008.

\bibitem{ekman1971constants}
Paul Ekman and Wallace~V Friesen.
\newblock Constants across cultures in the face and emotion.
\newblock {\em Journal of personality and social psychology}, 17(2):124, 1971.

\bibitem{wen2016discriminative}
Yandong Wen, Kaipeng Zhang, Zhifeng Li, and Yu~Qiao.
\newblock A discriminative feature learning approach for deep face recognition.
\newblock In {\em European conference on computer vision}, pages 499--515.
  Springer, 2016.

\bibitem{cai2018island}
Jie Cai, Zibo Meng, Ahmed~Shehab Khan, Zhiyuan Li, James O'Reilly, and Yan
  Tong.
\newblock Island loss for learning discriminative features in facial expression
  recognition.
\newblock In {\em 2018 13th IEEE International Conference on Automatic Face \&
  Gesture Recognition (FG 2018)}, pages 302--309. IEEE, 2018.

\bibitem{li2018facial}
Zhenghao Li, Song Wu, and Guoqiang Xiao.
\newblock Facial expression recognition by multi-scale cnn with regularized
  center loss.
\newblock In {\em 2018 24th International Conference on Pattern Recognition
  (ICPR)}, pages 3384--3389. IEEE, 2018.

\bibitem{farzaneh2021facial}
Amir~Hossein Farzaneh and Xiaojun Qi.
\newblock Facial expression recognition in the wild via deep attentive center
  loss.
\newblock In {\em Proceedings of the IEEE/CVF Winter Conference on Applications
  of Computer Vision}, pages 2402--2411, 2021.

\bibitem{fernandez2019feratt}
Pedro D~Marrero Fernandez, Fidel A~Guerrero Pena, Tsang~Ing Ren, and Alexandre
  Cunha.
\newblock Feratt: Facial expression recognition with attention net.
\newblock {\em arXiv preprint arXiv:1902.03284}, 3, 2019.

\bibitem{li2020attention}
Jing Li, Kan Jin, Dalin Zhou, Naoyuki Kubota, and Zhaojie Ju.
\newblock Attention mechanism-based cnn for facial expression recognition.
\newblock {\em Neurocomputing}, 411:340--350, 2020.

\bibitem{mase1991recognition}
Kenji Mase.
\newblock Recognition of facial expression from optical flow.
\newblock {\em IEICE TRANSACTIONS on Information and Systems},
  74(10):3474--3483, 1991.

\bibitem{wu2012survey}
Ting Wu, Siyao Fu, and Guosheng Yang.
\newblock Survey of the facial expression recognition research.
\newblock In {\em International Conference on Brain Inspired Cognitive
  Systems}, pages 392--402. Springer, 2012.

\bibitem{app13053259}
Luigi Bibbo’, Francesco Cotroneo, and Marley Vellasco.
\newblock Emotional health detection in har: New approach using ensemble snn.
\newblock {\em Applied Sciences}, 13(5), 2023.

\bibitem{s23052688}
Silvia Ceccacci, Andrea Generosi, Luca Giraldi, and Maura Mengoni.
\newblock Emotional valence from facial expression as an experience audit tool:
  An empirical study in the context of opera performance.
\newblock {\em Sensors}, 23(5), 2023.

\bibitem{dong2023recognizable}
Xiaoli Dong, Xin Ning, Jian Xu, Lina Yu, Weijun Li, and Liping Zhang.
\newblock A recognizable expression line portrait synthesis method in portrait
  rendering robot.
\newblock {\em IEEE Transactions on Computational Social Systems}, 2023.

\bibitem{rensink2000dynamic}
Ronald~A Rensink.
\newblock The dynamic representation of scenes.
\newblock {\em Visual cognition}, 7(1-3):17--42, 2000.

\bibitem{corbetta2002control}
Maurizio Corbetta and Gordon~L Shulman.
\newblock Control of goal-directed and stimulus-driven attention in the brain.
\newblock {\em Nature reviews neuroscience}, 3(3):201--215, 2002.

\bibitem{hu2018squeeze}
Jie Hu, Li~Shen, and Gang Sun.
\newblock Squeeze-and-excitation networks.
\newblock In {\em Proceedings of the IEEE conference on computer vision and
  pattern recognition}, pages 7132--7141, 2018.

\bibitem{qin2021fcanet}
Zequn Qin, Pengyi Zhang, Fei Wu, and Xi~Li.
\newblock Fcanet: Frequency channel attention networks.
\newblock In {\em Proceedings of the IEEE/CVF international conference on
  computer vision}, pages 783--792, 2021.

\bibitem{li2019spatial}
Xiang Li, Xiaolin Hu, and Jian Yang.
\newblock Spatial group-wise enhance: Improving semantic feature learning in
  convolutional networks.
\newblock {\em arXiv preprint arXiv:1905.09646}, 2019.

\bibitem{woo2018cbam}
Sanghyun Woo, Jongchan Park, Joon-Young Lee, and In~So Kweon.
\newblock Cbam: Convolutional block attention module.
\newblock In {\em Proceedings of the European conference on computer vision
  (ECCV)}, pages 3--19, 2018.

\bibitem{hou2021coordinate}
Qibin Hou, Daquan Zhou, and Jiashi Feng.
\newblock Coordinate attention for efficient mobile network design.
\newblock In {\em Proceedings of the IEEE/CVF conference on computer vision and
  pattern recognition}, pages 13713--13722, 2021.

\bibitem{misra2021rotate}
Diganta Misra, Trikay Nalamada, Ajay~Uppili Arasanipalai, and Qibin Hou.
\newblock Rotate to attend: Convolutional triplet attention module.
\newblock In {\em Proceedings of the IEEE/CVF Winter Conference on Applications
  of Computer Vision}, pages 3139--3148, 2021.

\bibitem{fu2019dual}
Jun Fu, Jing Liu, Haijie Tian, Yong Li, Yongjun Bao, Zhiwei Fang, and Hanqing
  Lu.
\newblock Dual attention network for scene segmentation.
\newblock In {\em Proceedings of the IEEE/CVF Conference on Computer Vision and
  Pattern Recognition}, pages 3146--3154, 2019.

\bibitem{liu2021swin}
Ze~Liu, Yutong Lin, Yue Cao, Han Hu, Yixuan Wei, Zheng Zhang, Stephen Lin, and
  Baining Guo.
\newblock Swin transformer: Hierarchical vision transformer using shifted
  windows.
\newblock {\em arXiv preprint arXiv:2103.14030}, 2021.

\bibitem{liu2022dab}
Shilong Liu, Feng Li, Hao Zhang, Xiao Yang, Xianbiao Qi, Hang Su, Jun Zhu, and
  Lei Zhang.
\newblock Dab-detr: Dynamic anchor boxes are better queries for detr.
\newblock {\em arXiv preprint arXiv:2201.12329}, 2022.

\bibitem{ding2022davit}
Mingyu Ding, Bin Xiao, Noel Codella, Ping Luo, Jingdong Wang, and Lu~Yuan.
\newblock Davit: Dual attention vision transformers.
\newblock In {\em Computer Vision--ECCV 2022: 17th European Conference, Tel
  Aviv, Israel, October 23--27, 2022, Proceedings, Part XXIV}, pages 74--92.
  Springer, 2022.

\bibitem{zhang2023vision}
Qiming Zhang, Jing Zhang, Yufei Xu, and Dacheng Tao.
\newblock Vision transformer with quadrangle attention.
\newblock {\em arXiv preprint arXiv:2303.15105}, 2023.

\bibitem{xie2019deep}
Siyue Xie, Haifeng Hu, and Yongbo Wu.
\newblock Deep multi-path convolutional neural network joint with salient
  region attention for facial expression recognition.
\newblock {\em Pattern recognition}, 92:177--191, 2019.

\bibitem{zhu2019discriminative}
Kangkang Zhu, Zhengyin Du, Weixin Li, Di~Huang, Yunhong Wang, and Liming Chen.
\newblock Discriminative attention-based convolutional neural network for 3d
  facial expression recognition.
\newblock In {\em 2019 14th IEEE International Conference on Automatic Face \&
  Gesture Recognition (FG 2019)}, pages 1--8. IEEE, 2019.

\bibitem{ning2021jwsaa}
Xin Ning, Ke~Gong, Weijun Li, and Liping Zhang.
\newblock Jwsaa: joint weak saliency and attention aware for person
  re-identification.
\newblock {\em Neurocomputing}, 453:801--811, 2021.

\bibitem{chen2021image}
Yuantao Chen, Linwu Liu, Volachith Phonevilay, Ke~Gu, Runlong Xia, Jingbo Xie,
  Qian Zhang, and Kai Yang.
\newblock Image super-resolution reconstruction based on feature map attention
  mechanism.
\newblock {\em Applied Intelligence}, 51:4367--4380, 2021.

\bibitem{wang2021dm3loc}
Duolin Wang, Zhaoyue Zhang, Yuexu Jiang, Ziting Mao, Dong Wang, Hao Lin, and
  Dong Xu.
\newblock Dm3loc: multi-label mrna subcellular localization prediction and
  analysis based on multi-head self-attention mechanism.
\newblock {\em Nucleic Acids Research}, 49(8):e46--e46, 2021.

\bibitem{hadsell2006dimensionality}
Raia Hadsell, Sumit Chopra, and Yann LeCun.
\newblock Dimensionality reduction by learning an invariant mapping.
\newblock In {\em 2006 IEEE Computer Society Conference on Computer Vision and
  Pattern Recognition (CVPR'06)}, volume~2, pages 1735--1742. IEEE, 2006.

\bibitem{liu2017sphereface}
Weiyang Liu, Yandong Wen, Zhiding Yu, Ming Li, Bhiksha Raj, and Le~Song.
\newblock Sphereface: Deep hypersphere embedding for face recognition.
\newblock In {\em Proceedings of the IEEE conference on computer vision and
  pattern recognition}, pages 212--220, 2017.

\bibitem{liu2017learning}
Yu~Liu, Hongyang Li, and Xiaogang Wang.
\newblock Learning deep features via congenerous cosine loss for person
  recognition.
\newblock {\em arXiv preprint arXiv:1702.06890}, 2017.

\bibitem{wang2018cosface}
Hao Wang, Yitong Wang, Zheng Zhou, Xing Ji, Dihong Gong, Jingchao Zhou, Zhifeng
  Li, and Wei Liu.
\newblock Cosface: Large margin cosine loss for deep face recognition.
\newblock In {\em Proceedings of the IEEE conference on computer vision and
  pattern recognition}, pages 5265--5274, 2018.

\bibitem{deng2019arcface}
Jiankang Deng, Jia Guo, Niannan Xue, and Stefanos Zafeiriou.
\newblock Arcface: Additive angular margin loss for deep face recognition.
\newblock In {\em Proceedings of the IEEE/CVF Conference on Computer Vision and
  Pattern Recognition}, pages 4690--4699, 2019.

\bibitem{farzaneh2020discriminant}
Xiaojun~Qi Farzaneh, Amir~Hossein.
\newblock Discriminant distribution-agnostic loss for facial expression
  recognition in the wild.
\newblock In {\em Proceedings of the IEEE/CVF Conference on Computer Vision and
  Pattern Recognition Workshops}, pages 406--407, 2020.

\bibitem{he2016deep}
Kaiming He, Xiangyu Zhang, Shaoqing Ren, and Jian Sun.
\newblock Deep residual learning for image recognition.
\newblock In {\em Proceedings of the IEEE conference on computer vision and
  pattern recognition}, pages 770--778, 2016.

\bibitem{dhall2012collecting}
Abhinav Dhall, Roland Goecke, Simon Lucey, and Tom Gedeon.
\newblock Collecting large, richly annotated facial-expression databases from
  movies.
\newblock {\em IEEE multimedia}, 19(03):34--41, 2012.

\bibitem{li2018reliable}
Shan Li and Weihong Deng.
\newblock Reliable crowdsourcing and deep locality-preserving learning for
  unconstrained facial expression recognition.
\newblock {\em IEEE Transactions on Image Processing}, 28(1):356--370, 2018.

\bibitem{6130508}
Abhinav Dhall, Roland Goecke, Simon Lucey, and Tom Gedeon.
\newblock Static facial expression analysis in tough conditions: Data,
  evaluation protocol and benchmark.
\newblock In {\em 2011 IEEE International Conference on Computer Vision
  Workshops (ICCV Workshops)}, pages 2106--2112, 2011.

\bibitem{deng2019retinaface}
Jiankang Deng, Jia Guo, Yuxiang Zhou, Jinke Yu, Irene Kotsia, and Stefanos
  Zafeiriou.
\newblock Retinaface: Single-stage dense face localisation in the wild.
\newblock {\em arXiv preprint arXiv:1905.00641}, 2019.

\bibitem{liu2019pose}
Yuanyuan Liu, Jiyao Peng, Jiabei Zeng, and Shiguang Shan.
\newblock Pose-adaptive hierarchical attention network for facial expression
  recognition.
\newblock {\em arXiv preprint arXiv:1905.10059}, 2019.

\bibitem{siqueira2020efficient}
Henrique Siqueira, Sven Magg, and Stefan Wermter.
\newblock Efficient facial feature learning with wide ensemble-based
  convolutional neural networks.
\newblock In {\em Proceedings of the AAAI conference on artificial
  intelligence}, volume~34, pages 5800--5809, 2020.

\bibitem{wang2020region}
Kai Wang, Xiaojiang Peng, Jianfei Yang, Debin Meng, and Yu~Qiao.
\newblock Region attention networks for pose and occlusion robust facial
  expression recognition.
\newblock {\em IEEE Transactions on Image Processing}, 29:4057--4069, 2020.

\bibitem{wang2020suppressing}
Kai Wang, Xiaojiang Peng, Jianfei Yang, Shijian Lu, and Yu~Qiao.
\newblock Suppressing uncertainties for large-scale facial expression
  recognition.
\newblock In {\em Proceedings of the IEEE/CVF Conference on Computer Vision and
  Pattern Recognition}, pages 6897--6906, 2020.

\bibitem{vo2020pyramid}
Thanh-Hung Vo, Guee-Sang Lee, Hyung-Jeong Yang, and Soo-Hyung Kim.
\newblock Pyramid with super resolution for in-the-wild facial expression
  recognition.
\newblock {\em IEEE Access}, 8:131988--132001, 2020.

\bibitem{zhao2021robust}
Zengqun Zhao, Qingshan Liu, and Feng Zhou.
\newblock Robust lightweight facial expression recognition network with label
  distribution training.
\newblock In {\em Proceedings of the AAAI Conference on Artificial
  Intelligence}, volume~35, pages 3510--3519, 2021.

\bibitem{savchenko2021facial}
Andrey~V Savchenko.
\newblock Facial expression and attributes recognition based on multi-task
  learning of lightweight neural networks.
\newblock {\em arXiv preprint arXiv:2103.17107}, 2021.

\bibitem{li2021mvit}
Hanting Li, Mingzhe Sui, Feng Zhao, Zhengjun Zha, and Feng Wu.
\newblock Mvit: Mask vision transformer for facial expression recognition in
  the wild.
\newblock {\em arXiv preprint arXiv:2106.04520}, 2021.

\bibitem{li2019separate}
Yingjian Li, Yao Lu, Jinxing Li, and Guangming Lu.
\newblock Separate loss for basic and compound facial expression recognition in
  the wild.
\newblock In {\em Asian Conference on Machine Learning}, pages 897--911. PMLR,
  2019.

\bibitem{chen2019facial}
Yuedong Chen, Jianfeng Wang, Shikai Chen, Zhongchao Shi, and Jianfei Cai.
\newblock Facial motion prior networks for facial expression recognition.
\newblock In {\em 2019 IEEE Visual Communications and Image Processing (VCIP)},
  pages 1--4. IEEE, 2019.

\bibitem{chen2020label}
Shikai Chen, Jianfeng Wang, Yuedong Chen, Zhongchao Shi, Xin Geng, and Yong
  Rui.
\newblock Label distribution learning on auxiliary label space graphs for
  facial expression recognition.
\newblock In {\em Proceedings of the IEEE/CVF Conference on Computer Vision and
  Pattern Recognition}, pages 13984--13993, 2020.

\bibitem{kollias2020deep}
Dimitrios Kollias, Shiyang Cheng, Evangelos Ververas, Irene Kotsia, and
  Stefanos Zafeiriou.
\newblock Deep neural network augmentation: Generating faces for affect
  analysis.
\newblock {\em International Journal of Computer Vision}, 128(5):1455--1484,
  2020.

\bibitem{ding2020occlusion}
Hui Ding, Peng Zhou, and Rama Chellappa.
\newblock Occlusion-adaptive deep network for robust facial expression
  recognition.
\newblock In {\em 2020 IEEE International Joint Conference on Biometrics
  (IJCB)}, pages 1--9. IEEE, 2020.

\bibitem{cai2021identity}
Jie Cai, Zibo Meng, Ahmed~Shehab Khan, James O’Reilly, Zhiyuan Li, Shizhong
  Han, and Yan Tong.
\newblock Identity-free facial expression recognition using conditional
  generative adversarial network.
\newblock In {\em 2021 IEEE International Conference on Image Processing
  (ICIP)}, pages 1344--1348. IEEE, 2021.

\bibitem{meng2017identity}
Zibo Meng, Ping Liu, Jie Cai, Shizhong Han, and Yan Tong.
\newblock Identity-aware convolutional neural network for facial expression
  recognition.
\newblock In {\em 2017 12th IEEE International Conference on Automatic Face \&
  Gesture Recognition (FG 2017)}, pages 558--565. IEEE, 2017.

\bibitem{yan2019cross}
Keyu Yan, Wenming Zheng, Tong Zhang, Yuan Zong, Chuangao Tang, Cheng Lu, and
  Zhen Cui.
\newblock Cross-domain facial expression recognition based on transductive deep
  transfer learning.
\newblock {\em IEEE Access}, 7:108906--108915, 2019.

\bibitem{aouayeb2021learning}
Mouath Aouayeb, Wassim Hamidouche, Catherine Soladie, Kidiyo Kpalma, and Renaud
  Seguier.
\newblock Learning vision transformer with squeeze and excitation for facial
  expression recognition.
\newblock {\em arXiv preprint arXiv:2107.03107}, 2021.

\bibitem{wu2021facecaps}
Fangyu Wu, Chaoyi Pang, and Bailing Zhang.
\newblock Facecaps for facial expression recognition.
\newblock {\em Computer Animation and Virtual Worlds}, page e2021, 2021.

\end{thebibliography}

\end{document}